\documentclass[conference,compsoc]{IEEEtran}
\usepackage{placeins}
\usepackage{amsmath,graphicx}
\usepackage{multirow}
\usepackage[font=footnotesize ,skip=0pt]{caption}

\begin{document}
		\nocite{*}
		
\title{Simplified Long Short-term Memory \\ Recurrent Neural Networks: part III}

\author{\IEEEauthorblockN{Atra Akandeh and Fathi M. Salem}
\IEEEauthorblockA{Circuits, Systems, and Neural Networks (CSANN) Laboratory \\
Computer Science and Engineering , Electrical and Computer Engineering \\
Michigan State University\\
East Lansing, Michigan 48864-1226\\
akandeha@msu.edu; salemf@msu.edu }
}

\maketitle

\begin{abstract}
This is part III of three-part work. In parts I and II, we have presented eight variants for simplified Long Short Term Memory (LSTM) recurrent neural networks (RNNs). It is noted that fast computation, specially in constrained computing resources, are an important factor in processing big time-sequence data. In this part III paper, we present and evaluate two new LSTM model variants which dramatically reduce the computational load while retaining comparable performance to the base (standard) LSTM RNNs. In these new variants, we impose (Hadamard) pointwise state multiplications in the cell-memory network in addition to the gating signal networks.
\end{abstract}


\IEEEpeerreviewmaketitle

\section{Introduction}
Nowadays Neural Networks play a great role in Information and Knowledge Engineering in diverse media forms including text, language, image, and video. Gated Recurrent Neural Networks have shown impressive performance in numerous applications in these domains [1-8].  We begin with the simple building block for clarity, namely the simple RNN. The simple RNN is is expressed using following equations:
\begin{equation}
	\begin{split}
		& h_t = \sigma(W_{hx} x_t + W_{hh} h_{t-1} + b_h) \\
		& y_t = W_{hy} h_t + b_y
	\end{split}
\end{equation}
The Gated RNNs, called Long Short-term Memory (LSTM) RNNs, were introduced in \cite{lstm}, by defining the concept of gating signals to control the flow of information [1-5]. A base (standard) LSTM model can be expressed as
\begin{equation}
	\begin{split}
		& i_t = \sigma_{in}(W_i x_t + U_i h_{t-1} + b_i) \\
		& f_t = \sigma_{in}(W_f x_t + U_f h_{t-1} + b_f) \\
		& o_t = \sigma_{in}(W_o x_t + U_o h_{t-1} + b_o) \\
		& \tilde{c_t} = \sigma(W_c x_t + U_c h_{t-1} + b_c) \\
		& c_t= f_t \odot c_{t-1} + i_t \odot \tilde{c_t}\\
		& h_t = o_t \odot \sigma(c_t) 
	\end{split}
\end{equation}
The first three equations express the three gating control signals. The three remaining equations express the main cell-memory network. In this part III paper, we shall apply parameter reductions to the main network! Only the state and the bias are candidates. We describe and evaluate two new simplified LSTM variants by uniformly reducing blocks of adaptive parameters in the gating mechanisms and also in main equation of the gated system.
\section{New Variants LSTM Models}
In part I and part II of this study, we introduced eight variants. In this part III, we present two new model variants. We seek to reduce the number of parameters and thus computational cost in this endeavor. 

\subsection{LSTM6}
This minimal model variant was introduced earlier and it is included here for baseline comparison reasons. Only constants has been selected for the gate equation, i.e there is no parameter associate with input, output and forget gate. The forget gate value must be less than one in absolute value for bounded-input-bounded-output (BIBO) stability \cite{salem2016_basic}.
\begin{equation}
	\begin{split}
		& i_t = 1.0 \\
		& f_t = 0.59 \\
		& o_t = 1 \\
		& \tilde{c_t} = \sigma(W_c x_t + U_c h_{t-1} + b_c) \\
		& c_t= f_t \odot c_{t-1} + i_t \odot \tilde{c_t}\\
		& h_t = o_t \odot \sigma(c_t) 
	\end{split}
\end{equation}
Note when the gate signal value is set to 1, this is, in practice, equivalent to eliminating the gate! The next two models perform nuances parameter reductions on the cell-body network equations. We figured using a numbering systems that start from 10 for ease for distinct referencing.

\subsection{LSTM10}
In this model, point-wise multiplication are applied to the hidden state and corresponding weights in the cell-body equations as well. We apply this modification not only to the gating equations but also to the main equation, i.e. matrix $U_c$ is replaced with vector $u_c$ for the pointwise multiplication.
\begin{equation}
	\begin{split}
		& i_t = \sigma_{in}(u_i \odot  h_{t-1}) \\
		& f_t = \sigma_{in}(u_f \odot  h_{t-1}) \\
		& o_t = \sigma_{in}(u_o \odot  h_{t-1}) \\
		& \tilde{c_t} = \sigma(W_c x_t + u_c \odot h_{t-1} + b_c) \\
		& c_t= f_t \odot  c_{t-1} + i_t \odot \tilde{c_t}\\
		& h_t = o_t \odot \sigma(c_t) 
	\end{split}
\end{equation}

\subsection{LSTM11}
This variant is similar to the LSTM10. However, it reinstates the biases in the gating signals. Mathematically, it is expressed as 
\begin{equation}
	\begin{split}
		& i_t = \sigma_{in}(u_i  \odot  h_{t-1} + b_i) \\
		& f_t = \sigma_{in}(u_f \odot  h_{t-1} + b_f) \\
		& o_t = \sigma_{in}(u_o \odot  h_{t-1} + b_o) \\
		& \tilde{c_t} = \sigma(W_c x_t + u_c \odot h_{t-1} + b_c) \\
		& c_t= f_t \odot c_{t-1} + i_t \odot \tilde{c_t}\\
		& h_t = o_t \odot \sigma(c_t) 
	\end{split}
\end{equation}

\begin{table}
	\caption{variants specifications.}
	\centering
	\begin{tabular}{| c | c | c |} 
		\hline
		variants & \# of parameters & times(s) per epoch \\ 
		\hline
		LSTM & 52610 & 30 \\
		\hline
		LSTM6 & 13910 & 12 \\
		\hline
		LSTM10 & 4310 & 18 \\
		\hline
		LSTM11 & 4610 & 19 \\
		\hline
	\end{tabular}
	\label{vs}
\end{table}

Table~\ref{vs} provides the total number of parameters and the comparative elapsed times per epoch corresponding to each variant.
\FloatBarrier
\section{Experiments and Discussion}
We have trained the variants on the benchmark MNIST dataset. The $28 \times 28$ image is passed to the network as row-wise sequences. Each network reads one row at a time and infer its decision after all rows have been read. In all cases, the variants have been trained using the Keras Library \cite{keras_rowwise}. Table~\ref{tab:a} summarizes the specification of the network architecture used. 

\begin{table}[h]
	\caption{Network specifications.}
	\centering
	\begin{tabular}{| c | c |} 
		\hline
		Input dimension & $28 \times 28 $ \\ 
		\hline
		Number of hidden units & 100 \\
		\hline
		Non-linear function & tanh, sigmoid, tanh \\
		\hline
		Output dimension & 10 \\
		\hline
		Non-linear function & softmax \\
		\hline
		Number of epochs & $100$ \\
		\hline
		Optimizer & RMprop \\
		\hline
		Batch size & 32 \\
		\hline
		Loss function & categorical cross-entropy \\
		\hline
	\end{tabular}
	\label{tab:a}
\end{table}

\subsection{Default $\eta$}
Initially, we picked $0.001$ for $\eta$. In the cases with $sigmoid$ or $tanh$ activation, all variants performed comparatively well. However, using the $relu$ activation caused model LSTM10 drop its accuracy performance to 52\%. Also accuracy of the base (standard) LSTM dropped after 50 epochs. The best test accuracy of the base LSTM is around 99\% and the test accuracy of LSTM10 and LSTM11 are respectively about 92\% and 95\% using $tanh$. Other cases are summarized in Table~\ref{def}. We explored a range of $\eta$ for $sigmoid$ and $tanh$ in which variants LSTM10 and LSTM11 can become competitive within the 100 epochs. We also explored a valid range of $\eta$ for $relu$. 

\begin{figure}[!htb]
	\centering
	\setlength{\belowcaptionskip}{-15pt}
	\includegraphics[trim={0 0 0 0},clip,scale=0.42]{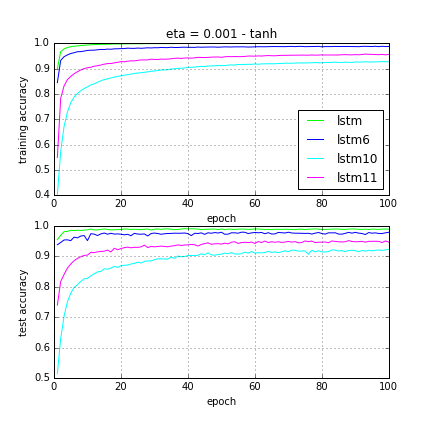}
	\caption{Training \& Test accuracy, $\sigma=tanh , \eta=1\mathrm{e}{-3}$}
	\label{fig:fig1}
\end{figure}

\begin{figure}[!htb]
	\centering
	\includegraphics[trim={0 0 0 0},clip,scale=0.42]{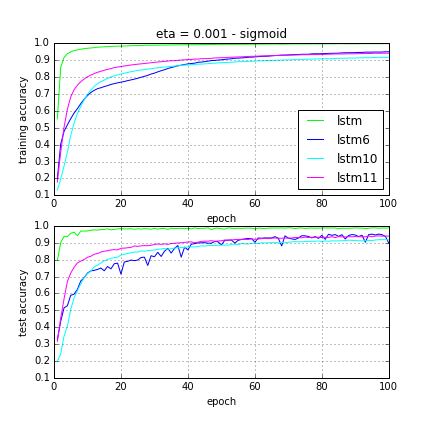}
	\caption{Training \& Test accuracy, $\sigma=sigmoid , \eta=1\mathrm{e}{-3}$}
	\label{fig:fig2}
\end{figure}

\begin{figure}[!htb]
	\centering
	\includegraphics[trim={0 0 0 0},clip,scale=0.42]{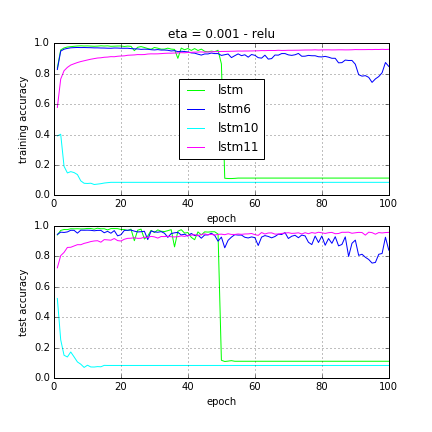}
	\caption{Training \& Test accuracy, $\sigma=relu , \eta=1\mathrm{e}{-3}$}
	\label{fig:fig3}
\end{figure}

\begin{table}[!htb]
	\caption{Best results obtained by $\eta = 0.001$.}
	\centering
	\begin{tabular}{ |c|c|c|c|c| }
		\cline{3-5}
		\multicolumn{1}{c}{}
		& & $tanh$ & $sigmoid$ & $relu$ \\
		\hline
		\multirow{2}{*}{LSTM} & train &  1.000 &  0.9972 &  0.9829 \\ 
		& test &  0.9909 &  0.9880 &  0.9843 \\ 
		\hline
		\multirow{2}{*}{LSTM6} & train &  0.9879 &  0.9495 &  0.9719 \\ 
		& test &   0.9792 &  0.9513 &  0.9720 \\ 
		\hline
		\multirow{2}{*}{LSTM10} & train &  0.9273 &  0.9168 &  0.4018 \\ 
		& test &  0.9225 &  0.9184 &  0.5226 \\ 
		\hline
		\multirow{2}{*}{LSTM11} & train &  0.9573 &  0.9407 &  0.9597 \\ 
		& test &  0.9514 &  0.9403 &  0.9582 \\ 
		\hline
	\end{tabular}
	\label{def}
\end{table}

\subsection{Searching for best $\eta$}
We increased $\eta$ from $0.001$ to $0.005$ in increments of $0.001$. This led into an increase in test accuracy of model LSTM11 and model LSTM12, yielding values 95.31\% and 93.56\% respectively for the $tanh$ case. As expected, LSTM10 with the $relu$ activation failed progressively comparing to smaller $\eta$ values.  The training and test accuracy of these new $\eta$ values are shown in Table~\ref{lstm10-et} and Table~\ref{lstm11-et}.
\begin{figure}[!htb]
	\centering
	\includegraphics[trim={0 0 0 0},clip,scale=0.42]{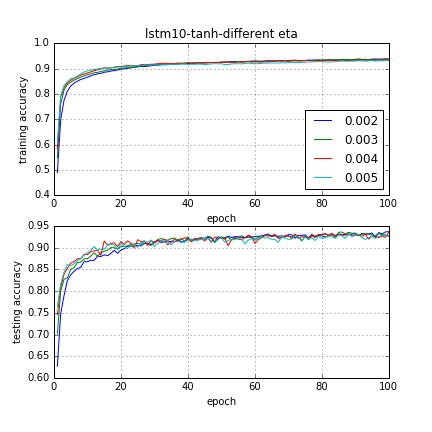}
	\caption{Training \& Test accuracy of different $\eta$, lstm10, $tanh$}
	\label{fig:fig4}
\end{figure}

\begin{figure}[!htb]
	\centering
	\includegraphics[trim={0 0 0 0},clip,scale=0.42]{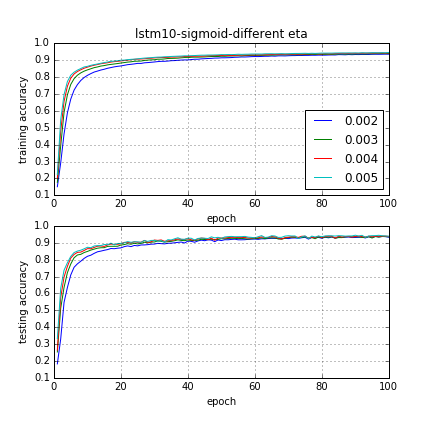}
	\caption{Training \& Test accuracy of different $\eta$, lstm10, $sigmoid$}
	\label{fig:fig5}
\end{figure}

\begin{figure}[!htb]
	\centering
	\includegraphics[trim={0 0 0 0},clip,scale=0.42]{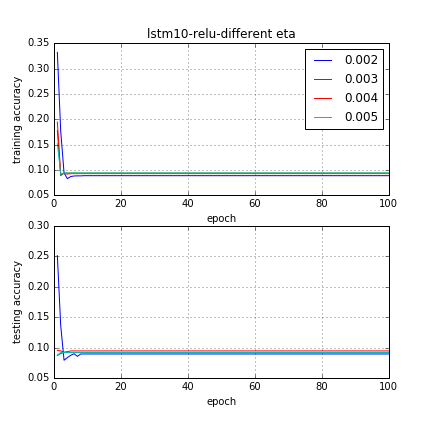}
	\caption{Training \& Test accuracy of different $\eta$, lstm10, $relu$}
	\label{fig:fig6}
\end{figure}

\begin{figure}[!htb]
	\centering
	\includegraphics[trim={0 0 0 0},clip,scale=0.42]{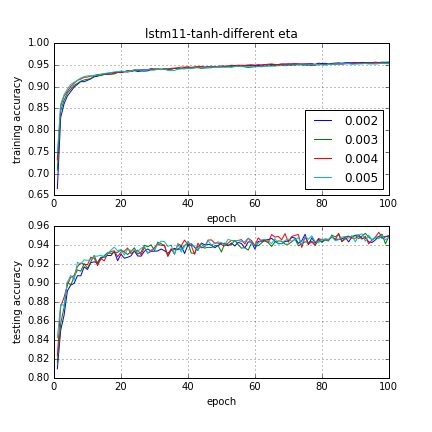}
	\caption{Training \& Test accuracy of different $\eta$, lstm11, $tanh$}
	\label{fig:fig7}
\end{figure}

\begin{figure}[!htb]
	\centering
	\includegraphics[trim={0 0 0 0},clip,scale=0.42]{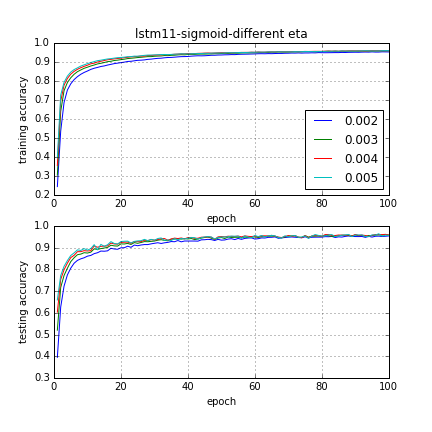}
	\caption{Training \& Test accuracy of different $\eta$, lstm11, $sigmoid$}
	\label{fig:fig8}
\end{figure}

\begin{figure}[!htb]
	\centering
	\includegraphics[trim={0 0 0 0},clip,scale=0.42]{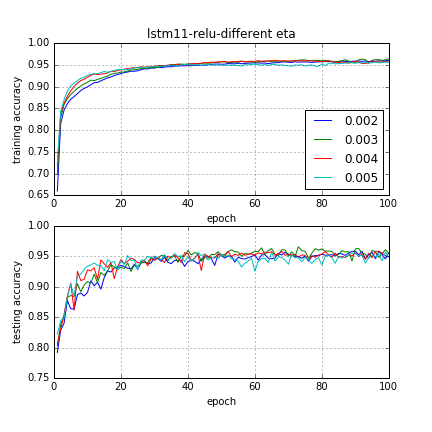}
	\caption{Training \& Test accuracy of different $\eta$, lstm11, $relu$}
	\label{fig:fig9}
\end{figure}

\begin{table}[!htb]
	\caption{Best results obtained by LSTM10.}
	\centering
	\begin{tabular}{ |c|c|c|c|c| }
		\cline{3-5}
		\multicolumn{1}{c}{}
		& & $tanh$ & $sigmoid$ & $relu$ \\
		\hline
		\multirow{2}{*}{$\eta=0.002$} & train &  0.9376 &  0.9354 &  0.3319 \\ 
		& test &  0.9366 &  0.9376 &  0.2510 \\ 
		\hline
		\multirow{2}{*}{$\eta=0.003$} & train &  0.9388 &  0.9389 &  0.1777 \\ 
		& test &   0.9357 &  0.9367 &  0.0919 \\ 
		\hline
		\multirow{2}{*}{$\eta=0.004$} & train &  0.9348 &  0.9428 &  0.1946 \\ 
		& test &  0.9350 &  0.9392 &  0.0954 \\ 
		\hline
		\multirow{2}{*}{$\eta=0.005$} & train &  0.9317 &  0.9453 &  0.1519 \\ 
		& test &  0.9318 &  0.9444 &  0.0919 \\ 
		\hline
	\end{tabular}
	\label{lstm10-et}
\end{table}

\begin{table}[!htb]
	\caption{Best results obtained by LSTM11.}
	\centering
	\begin{tabular}{ |c|c|c|c|c| }
		\cline{3-5}
		\multicolumn{1}{c}{}
		& & $tanh$ & $sigmoid$ & $relu$ \\
		\hline
		\multirow{2}{*}{$\eta=0.002$} & train &  0.9566 &  0.9546 &  0.9602 \\ 
		& test &  0.9511 &  0.9534 &  0.9583 \\ 
		\hline
		\multirow{2}{*}{$\eta=0.003$} & train &  0.9557 &  0.9601 &  0.9637 \\ 
		& test &   0.9521 &  0.9598 &  0.9656 \\ 
		\hline
		\multirow{2}{*}{$\eta=0.004$} & train &  0.9552 &  0.9608 &  0.9607 \\ 
		& test &  0.9531 &  0.9608 &  0.9582 \\ 
		\hline
		\multirow{2}{*}{$\eta=0.005$} & train &  0.9539 &  0.9611 &  0.9565 \\ 
		& test &  0.9516 &  0.9635 &  0.9569 \\ 
		\hline
	\end{tabular}
	\label{lstm11-et}
\end{table}

\subsection{Finding $\eta$ for LSTM10 relu }
We have explored a range of $\eta$ for LSTM10 with $relu$ activation to improve its performance. However, the effort was not successful. Increasing $\eta$ from $2e-6$ to $1e-5$ leads to an increase in accuracy with value of 53.13\%.  For $\eta$ less than $1\mathrm{e}{-5}$, the plots have increasing trend. However, after this point, the accuracy starts to drop after a number of epochs depending on the value of $\eta$.

\begin{figure}[!htb]
	\centering
	\includegraphics[trim={0 0 0 0},clip,scale=0.42]{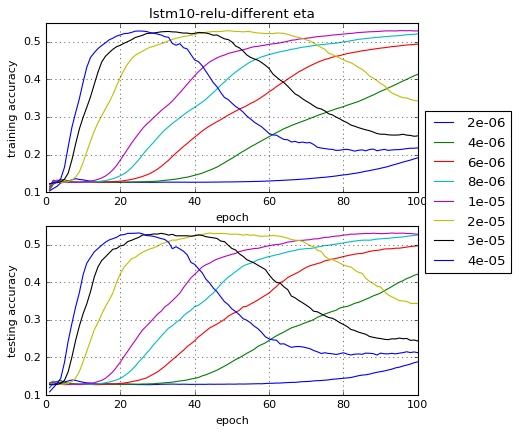}
	\caption{Training \& Test accuracy of different $\eta$, lstm10, $relu$}
	\label{fig:fig10}
\end{figure}

\section{Conclusion}
In this study, we have described and evaluated two new reduced variants of LSTM model. We call these new models LSTM10 and LSTM11. These models have been examined and evaluated on the MNIST dataset with different activations and different learning rate $\eta$ values. In our part I and part II, we considered variants to the base LSTM by removing weights/biases from the gating equations only. In this study, we have reduced weights even within the main cell-memory equation of the model by converting a weight matrix to a vector and replace regular multiplication with (Hadamard) pointwise multiplication. The only difference between model LSTM10 and LSTM11 is that the latter retain the bias term in the gating equations. LSTM 6 is equivalent to the so-called basic recurrent neural network (bRNN), since all gating equation have been replaced by a fixed constant-- see \cite{salem2016_basic}. It has been found that all of variants, except model LSTM10 when using the activation $relu$, are comparable to a (standard) base LSTM RNN. We anticipate that further case studies and experiments would serve to fine-tune these findings. 


%
\section*{Acknowledgment}
This work was supported in part by the National Science Foundation under grant No. ECCS-1549517.

\small{
	\bibliographystyle{ieee}
	\bibliography{egbib}
}

\end{document}